\documentclass[journal,10pt]{IEEEtran}

\usepackage{amsmath, amssymb}
\usepackage{graphicx}
\usepackage[pagebackref=true,breaklinks=true,letterpaper=true,colorlinks,bookmarks=false]{hyperref}
\usepackage{multirow}

\usepackage{authblk}

\usepackage{parskip}

\usepackage{subcaption}

\usepackage{bm}

\usepackage{balance}

\usepackage[dvipsnames]{xcolor}
\usepackage{tikz}
\usepackage{pgfplots}

\newif\ifarxiv
\arxivtrue

\ifarxiv
\makeatletter
\def\ps@IEEEtitlepagestyle{%
  \def\@oddfoot{\copyrightnotice}%
  \def\@evenfoot{}%
}
\def\copyrightnotice{%
  {\footnotesize Accepted for publication in IEEE Transactions on Image Processing: \url{https://doi.org/10.1109/TIP.2018.2867270}\hfill}% <--- Change here
  \gdef\copyrightnotice{}% just in case
}
\fi

\begin{document}

	\title
	{
		Learning Local Descriptors by Optimizing the Keypoint-Correspondence Criterion:
			Applications to Face Matching, Learning from Unlabeled Videos and 3D-Shape Retrieval
	}

	\author[$\dagger$]{Nenad Marku\v{s}\thanks{\texttt{nenad.markus@fer.hr}}}
	\author[$\dagger$]{Igor S. Pand\v{z}i\'{c}}
	\author[$\ddagger$]{J\"orgen Ahlberg}
	\affil[$\dagger$]
	{
		\small
		University of Zagreb,
		Faculty of Electrical Engineering and Computing,
		Unska 3, 10000 Zagreb, Croatia
	}
	\affil[$\ddagger$]
	{
		\small
		Computer Vision Laboratory,
		Dept.~of Electrical Engineering,
		Link\"{o}ping University,
		SE-581 83 Link\"{o}ping, Sweden
	}

	\date{}

	\maketitle

	\begin{abstract}
		Current best local descriptors are learned on a large dataset of matching and non-matching keypoint pairs.
		However, data of this kind is not always available since detailed keypoint correspondences can be hard to establish.
		On the other hand, we can often obtain labels for pairs of keypoint bags.
		For example, keypoint bags extracted from two images of the same object under different views form a matching pair, and keypoint bags extracted from images of different objects form a non-matching pair.
		On average, matching pairs should contain more corresponding keypoints than non-matching pairs.
		We describe an end-to-end differentiable architecture that enables the learning of local keypoint descriptors from such weakly-labeled data.
		Additionally, we discuss how to improve the method by incorporating the procedure of mining hard negatives.
		We also show how can our approach be used to learn convolutional features from unlabeled video signals and 3D models.
	\end{abstract}

	\section{Introduction}
		Local descriptors are a widely used tool in computer vision and pattern recognition.
		Some example applications include object/scene recognition and retrieval \cite{visualwords,fisherkernel,vlad}, face verification \cite{fisherfaces,localfacerec}, face alignment \cite{sdm}, image stitching \cite{recognisingpanoramas}, 3D shape estimation \cite{shapefrom2dlandmarks}, 3D model retrieval/matching \cite{hybridbow,cov3d} and visual SLAM \cite{orbslam}.
		However, despite years of research, there is still room for improvement, as confirmed by recent results based on convolutional neural networks \cite{zagoruyko,matchnet,convdesc,pnnet,LIFT,l2net,hardnet}.
		Also, we view the research in local descriptors complementary to keypoint detection research, which is still an active area of computer vision (see, for example \cite{tilde}).
		There is also research that aims to improve descriptor-matching techniques \cite{gms}.

		Discriminative local descriptors can be learned from annotated keypoint correspondences.
		This can be used to form a set of matching and non-matching keypoint pairs:
		\begin{equation}\label{eq:kppairs}
			\mathcal{D}_{KP}=
			\{(k_{i1}, k_{i2}, l_i)\}_{i=1}^N
			.
		\end{equation}
		The label $l_i\in\{+1, -1\}$ indicates whether keypoints $k_{i1}$ and $k_{i2}$ form a matching or a non-matching pair.
		See \cite{brown,binboost,rfd,zagoruyko,matchnet,convdesc,LIFT,l2net,hardnet} for some recent examples of descriptor-learning methods that use data in this form.
		Another possibility is to form a set of keypoint triplets:
		\begin{equation}\label{eq:kptriplets}
			\mathcal{D}_{KT}=
			\{(k_i, k_i^+, k_i^-)\}_{i=1}^N
			,
		\end{equation}
		where $k_i$ and $k_i^+$ match and $k_i$ and $k_i^-$ do not.
		For example, Balntas et al. \cite{pnnet} use data in this form in their method.
		The standard dataset for learning and benchmarking various image keypoint descriptors was introduced by Brown et al. \cite{brown}. %: \url{http://www.cs.ubc.ca/~mbrown/patchdata/patchdata.html}.
		It contains around $1.5$M patches cropped around difference of Gaussians (DoG) keypoints \cite{sift} obtained from multiple views of three different scenes: the Notre Dame Cathedral, the Statue of Liberty and the Yosemite Half Dome.
		High-quality keypoint labels were obtained with a multi-view stereo algorithm \cite{multiviewstereo}.
		This makes the dataset reliable both for learning local image-patch descriptors from "handcrafted" features \cite{cvx,binboost,rfd} and large models based on convolutional neural networks \cite{osendorfer,zagoruyko,matchnet,fracking,convdesc,pnnet,l2net,hardnet}
		(the Siamese-network framework of Hadsell et al. \cite{siamesenet}).
		However, in the general case, this kind of data is relatively hard to obtain, even more so for non-image data (e.g., 3D models, depth maps, voxel data, video signals, etc.).

		Instead of having a dataset with individual keypoint correspondences (which lead to dataset types \eqref{eq:kppairs} and \eqref{eq:kptriplets}) for learning local descriptors as in most prior work, we assume a set of labeled \emph{bags of keypoints}
		(here we intentionally use the terminology from multiple instance learning \cite{MIL} as our ideas are closely related with the field).
		We denote this weakly-labeled dataset as
		\begin{equation}\label{eq:bagtriplets}
			\mathcal{D}_{BT}=
			\{(K_{i}, K_i^+, K_i^-)\}_{i=1}^N
			,
		\end{equation}
		where bags $K_i$ and $K_i^+$ form a matching pair, bags $K_i$ and $K_i^-$ form a non-matching pair, and each bag is a set of $n$ keypoints, $K=\{k_1, k_2, \ldots, k_n\}$.
		Data of this kind is relatively easy to generate:
		keypoint bags extracted from two images of the same object under different views form a matching pair.
		These bags can be used together with a keypoint bag extracted from an image of some unrelated object to form a triplet from Equation \eqref{eq:bagtriplets}.
		See Figure \ref{fig:teaser} for an illustration.
		\input{teaser.fig}

		In this paper we expand our previous work \cite{wlrnicpr}.
		There we introduced a method for learning local descriptors from weakly-labeled data and provided preliminary experimental verification of its usefulness.
		Here we develop the obtained results further and show how to improve the method by incorporating hard-negative mining.
		We provide strong evidence that it is important to tune the descriptor for the visual appearance of the dataset.
		This makes our method particularly useful since it enables learning from weaker annotations than traditional descriptor-learning techniques
		(this is potentially cheaper and more time-efficient).
		Furthermore, we show how the method can be used in biometric systems and introduce two novel methods for learning local descriptors from unlabeled videos and 3D shapes.
		We also compare descriptors learned with our method to competing ones on an independent benchmark \cite{hpatches}.

	\section{Related work}
		We already mentioned a large body of work in local image descriptors and we will not repeat these standard approaches.

		We would like to mention the work of Paulin et al. \cite{kernelconv} since they are also motivated to obtain discriminative local descriptors by means that do not require strongly-labeled data (equations \eqref{eq:kppairs} and \eqref{eq:kptriplets}).
		To achieve their goal, they adapt the convolutional kernel-network approach, which is an unsupervised framework for learning convolutional architectures \cite{kernelconv0}.

		The learning procedure we propose in the next section is related to the one by Arandjelovi\'{c} et al. \cite{netvlad}, as they also propose to learn descriptors from weakly-labeled data.
		Unlike us, they do not focus on local descriptors and learn whole image representations instead.
		Also, they derive their learning procedure from a different perspective: we are concerned with local image correspondences and how to find them, and they focus on learning a global descriptor for image retrieval.
		It is not clear how well would their system work in finding local correspondences between two images.
		Also, we learn our descriptors \textit{directly} for comparisons with $L_2$ distance.

		There are also approaches that learn large, convolutional architectures to directly find correspondences between images \cite{warpnet,universalcorr} or estimate the optical flow \cite{flownet}.
		These approaches require a large dataset of annotated correspondences during training.
		This separates their work from ours since we aim to learn descriptors from weaker annotations.

	\section{Method}\label{sec:core}
		We study how to learn the parameters of a descriptor-extraction process that transforms a local neighborhood of a keypoint (e.g, a patch extracted around a distinctive corner within an image) into a short vector in such a way that similar keypoints are "close" and dissimilar keypoints are "far".
		Two attractive properties of such representations are low memory requirements and fast matching times.
		Unlike most prior work, our learning method exploits the information in weakly-labeled data to achieve mentioned goals.

		In this paper, we denote the descriptor-extraction process as $e$ (this is basically a number of predefined computational steps).
		For example, in our experiments, $e$ is a convolutional neural network (see Table \ref{tbl:nn} for its architecture) that maps a $32\times 32$ local image patch into a vector.
		We denote the parameters of $e$ as $\theta_e$.
		Here we describe an effective procedure for learning $\theta_e$ from the training data given by Equation \eqref{eq:bagtriplets}.
		First, we define that two keypoints match if the $L_2$ distance between their signatures (extracted by $e$) is less than or equal to some threshold $\tau\in\mathbb{R}$.
		This threshold is a parameter of the learning process and we specify some recommended values later in the text.
		Next, we define a \emph{matching score} between two bags of keypoints (both of size $n$), $K_1$ and $K_2$, as
		\begin{equation}\label{eq:bagsim}
			S_{e, \tau}(K_1, K_2)=
			\frac{m_{e, \tau}(K_1, K_2)}{n}
			,
		\end{equation}
		where $m_{e, \tau}(K_1, K_2)$ is the number of keypoints from $K_1$ that have a matching keypoint in $K_2$ for the descriptor extractor $e$ and threshold $\tau$.
		Optimal matching could be computed with the Hungarian algorithm in $O(n^3)$ time.
		However, this is too slow in our case and we use the following $O(n^2)$ approximation
		(inspired by the "sum-max" match kernel from \cite{summax}):
		\begin{equation}\label{eq:bagmatch}
			m_{e, \tau}(K_1, K_2)=
			\sum_{i=1}^n \left[ \min_{j=1}^n d_{ij}^2 \leq \tau \right]
			,
		\end{equation}
		where $[\;\cdot\;]$ represents the indicator function\footnote{$[p]=1$ if the proposition $p$ is true and $[p]=0$ otherwise.} and $d_{ij}$ is the Euclidean distance between descriptors of $k_i\in K_1$ and $k_j\in K_2$, i.e.,
		\begin{equation}
			d_{ij} =
			|| e(k_i) - e(k_j) ||_2
			.
		\end{equation}
		We want high $S_{e, \tau}$ for matching bags and low $S_{e, \tau}$ for non-matching bags.
		Thus, a suitable loss for parameter learning is
		\begin{equation}\label{eq:loss}
			L=
			\sum\frac{S_{e, \tau}(K, K^-)+\frac{1}{n}}{S_{e, \tau}(K, K^+)+\frac{1}{n}}
			,
		\end{equation}
		where the summation goes over $(K, K^+, K^-)\in\mathcal{D}_{BT}$ (Equation \eqref{eq:bagtriplets}) and $+\frac{1}{n}$ is included for numerical stability.
		However, since $S_{e, \tau}$ is not continuous, we cannot apply the standard gradient-based learning techniques.
		Thus, we resort to the following approximation of the function $[x\leq\tau]$ for $x\in\mathbb{R}$:
		\begin{equation}\label{eq:smooth}
			[x\leq\tau]\approx
			\frac{1}{1 + \exp(\beta(x-\tau))}
			,
		\end{equation}
		where the parameter $\beta$ regulates the "strength" of the approximation.
		Since the loss function $L$ is now differentiable, the parameters $\theta_e$ can be tuned with standard backpropagation-based methods:
		we approximate the solution with a local minimum to which the learning converges and experimentally show that this leads to good results.

		To simplify the implementation, we require that the extractor outputs descriptors of unit length: $||e(k_i)||_2=||e(k_j)||_2=1$.
		Notice that in this scenario
		\begin{equation}\label{eq:edist}
			d_{ij}^2=
			|| e(k_i) - e(k_j) ||_2^2 =
			2 - 2e(k_i)^Te(k_j)
		\end{equation}
		and the matching score function $S_{e, \tau}$ (Equation \eqref{eq:bagsim}) depends only on the matrix $\mathbf{S}\in\mathbb{R}^{n\times n}$ computed as
		\begin{equation}
			\mathbf{S}=
			\mathbf{E}_1\mathbf{E}_2^T
			,
		\end{equation}
		where the rows of matrices $\mathbf{E}_1$ and $\mathbf{E}_2$ contain descriptors extracted with the extractor $e$ from keypoints in $K_1$ and $K_2$.
		The backpropagation expressions are quite elegant in this setting:
		\begin{equation}\begin{aligned}
			\frac{\partial S_{e, \tau}(K_1, K_2)}{\partial \mathbf{E}_1}&=\frac{\partial S_{e, \tau}(K_1, K_2)}{\partial \mathbf{S}}\cdot\mathbf{E}_2	\\
			\frac{\partial S_{e, \tau}(K_1, K_2)}{\partial \mathbf{E}_2}&=\left(\frac{\partial S_{e, \tau}(K_1, K_2)}{\partial \mathbf{S}}\right)^T\cdot\mathbf{E}_1
		\end{aligned}\end{equation}
		where $\partial S_{e, \tau}(K_1, K_2)/\partial \mathbf{S}$ is straightforward to compute because $S_{e, \tau}$ contains only the standard components usually used in neural networks (see the definition, Equation \eqref{eq:bagsim}).
		The proposed computational steps can be implemented very efficiently in a few hundred lines of Torch code.
		Another advantage of unit-length descriptors is that this simplifies the selection of the threshold $\tau$: the Euclidean distance between two descriptors falls in the $[0, 2]$ interval (Equation \eqref{eq:edist}).

		We refer to the combination of equations \eqref{eq:bagsim}--\eqref{eq:smooth} as
		\textbf{Smoothed Keypoint-mAtching Ratio} (SKAR).
		We abbreviate the descriptors learned by propagating the gradient through this loss as SKAR descriptors.
		This notation is used in tables and graphs throughout the experimental part of the paper.

		\textbf{Hard-negative mining}.
		Some descriptor-learning methods (e.g., \cite{convdesc,l2net,hardnet}) incorporate a mechanism of finding the so-called hard negatives:
		non-matching patches that look sufficiently similar that the descriptor tends to confuse them for matching patches.
		The hypothesis is that the discriminative power of the descriptor increases when hard negatives are included in the learning process.
		Our method mines hard negatives when computing the similarity between $K$ and $K^-$ (the $\min$ operator from Equation \eqref{eq:bagmatch} takes care of this).
		However, $K^-$ contains only the keypoints extracted from a single non-matching image.
		The nature of annotated data is most of the time such that it is possible to generate a large number of negative bags for each $K$.
		This follows from the same reasoning that is used when mining hard negatives for learning descriptors in a strongly supervised manner.
		We propose to merge several negative bags into an \textit{augmented negative bag}:
		\begin{equation}\label{eq:hardnegs}
			K^-_*=
			\bigcup_j K^-_j
			, 
		\end{equation}
		where each $K^-_j$ is one of the non-matching bags to $K$.
		In practice, the union in Equation \eqref{eq:hardnegs} goes only over a random subset of all possible negative bags due to computational and storage reasons.
		We conjecture that using $(K, K^+, K^-_*)$ in the learning process described in this section can improve the matching performance of the descriptor:
		since $K^-_*$ is larger than $K^-$, the $\min$ operator from Equation \eqref{eq:bagmatch} can extract "harder" non-matching keypoints.
		This hypothesis is investigated in sections \ref{sec:weakly-labeled} and \ref{sec:hpatches}.

		The following sections describe experiments which show that the proposed learning procedure leads to good results with various diverse keypoint extractors
		(SIFT/DoG \cite{sift}, SURF \cite{surf}, ORB \cite{orb,FAST})
		and in several applications (image retrieval/matching, face verification, learning from unlabeled video signals, 3D-shape recognition).

	\section{Learning from weakly-labeled data}\label{sec:weakly-labeled}
		The experiments in this section complement the previously presented ones \cite{wlrnicpr}:
		we repeat the training and validation on a significantly larger number of patches and compare to recent state-of-the-art descriptors.

		We use the following datasets for our initial experiments:
		\begin{itemize}
			\item
				UKB \cite{ukb} ($2500$ objects, $4$ views each);
			\item
				ZuBuD \cite{zubud} ($200$ buildings, $5$ images each);
			\item
				INRIA Holidays \cite{holidays}
				(approximately $1500$ images of $500$ different scenes).
		\end{itemize}
		Each image is transformed into a bag of patches by running a keypoint detector over it.
		This sets up a basis for experimental comparison between different descriptors since we always use the same keypoints
		(location, size\footnote{A patch of a fixed size around the keypoint is resampled to $32\times 32$ or $64\times 64$ pixels, depending on the requirements of the descriptor-extraction process.} and orientation).
		We extract approximately $5$ times more keypoints per image than in our previous paper \cite{wlrnicpr}.

		The rest of this section is partitioned into four subsections.
		The SKAR learning process and parameters are described in the next subsection.
		The last three subsections describe retrieval-based experiments that compare SKAR descriptors to the state of the art.

		\subsection{Learning convolutional features with our method}\label{sec:learning}
			We use a similar descriptor extractor $e$ as in our previous paper \cite{wlrnicpr}.
			The architecture is specified in Table \ref{tbl:nn}.
			\begin{table}
				\resizebox{\columnwidth}{!}
				{
				\begin{tabular}{|c||c|c|c|c|}
					\hline
					Conv. layer	&	1	&	2	&	3	&	4	\\
					\hline
					\hline
					Filter size	&	$3\times 3$	&	$4\times 4$	&	$3\times 3$	&	$1\times 1$	\\
					\hline
					Stride	&	1	&	2	&	1	&	1	\\
					\hline
					Output channels	&	32	&	64	&	128	&	32	\\
					\hline
					Activation function	&	ReLU	&	ReLU	&	None	&	None	\\
					\hline
					Max pooling?	& No	&	No	&	Yes, $2\times2$	&	No	\\
					\hline
				\end{tabular}
				}
				\caption
				{
					Our descriptor extractor is a convolutional network that maps a $32\times 32$ RGB patch into a vector of fixed size.
					It consists of four convolutional layers (given in table above), a fully connected layer that maps the output of the last convolutional layer to $128$ neurons and a final $L_2$ normalization module (i.e., the output vector has unit length).
					The network has around $250$k parameters.
				}
				\label{tbl:nn}
			\end{table}
			Note that other differentiable architectures could be used as well.

			To generate the training data for our method, we partition the UKB dataset into two subsets.
			The larger subset contains $2200$ objects and is used to sample keypoint bag triplets \eqref{eq:bagtriplets}.
			This subset is used for learning and the rest of the UKB dataset ($300$ objects) is used for validation and testing.

			We learn $7$ SKAR descriptors on the allotted UKB partition.
			Each is trained on a different combination of keypoint types and number of negative bags.
			See Table \ref{tbl:skardescs} for details.
			\begin{table}
				\resizebox{\columnwidth}{!}
				{
				\begin{tabular}{|c||c|c|c|c|c|c|c|}
					\hline
					SKAR	&	O-1	&	O-6	&	O-12	&	S-1	&	S-6	&	S-12	&	S+O-12	\\
					\hline
					O&500&500&500&0&0&0&500\\
					\hline
					S&0&0&0&500&500&500&500\\
					\hline
					$K^-_*$ size&1&6&12&1&6&12&12\\
					\hline
				\end{tabular}
				}
				\caption
				{
					The nomenclature used for our SKAR descriptors learned on the UKB \cite{ukb} training partition (first row).
					Second and third row show the average number of ORB \cite{orb} and SIFT (DoG) \cite{sift} keypoints extracted per image during training, respectively.
					The last row shows how many negative bags were used to form $K^-_*$ (Equation \eqref{eq:hardnegs}).
				}
				\label{tbl:skardescs}
			\end{table}
			We set $\beta=20$ and $\tau=0.8$ (see Section \ref{sec:core} for their meaning), i.e., to same values as in our previous paper \cite{wlrnicpr}.
			During each training iteration, the loss $L$ (Equation \eqref{eq:loss}) is approximated by a minibatch of $32$ triplets $(K, K^+, K^-_*)$ and the parameters are slowly tuned with \texttt{rmsprop} (its learning rate is fixed to $10^{-4}$).
			The whole training process consists of approximately $10\;000$ such minibatches.
			This takes around one day for $K^-_*$ of size $12$ on a modern machine with $4$ GPUs.

		\subsection{Matching-based retrieval}
			As in our previous work \cite{wlrnicpr}, we implement a simple visual search engine.
			The retrieval is based on the number of matching descriptors between the query image and each of the other images from the dataset:
			the image with more matches is assigned a better rank.
			Each image of the dataset is used as a query once.
			We use the ratio criterion, proposed by Lowe \cite{sift}, to determine whether two descriptors match.
			This consists of the following steps for each descriptor $d$ from the query bag:
			\begin{enumerate}
				\item
					in the database bag find two $L_2$ closest descriptors (denote them as $d_{NN(1)}$ and $d_{NN(2)}$);
				\item
					compute the ratio $r$: $r=\frac{||d-d_{NN(1)}||_2}{||d-d_{NN(2)}||_2}$;
				\item
					if $r<\tau$, the descriptors $d$ and $d_{NN(1)}$ match.
			\end{enumerate}
			The threshold $\tau$ is selected separately from the set $\{0.7, 0.75, 0.8, 0.85, 0.9\}$ for each descriptor and each dataset to produce the best retrieval results.
			This approximates the raw discriminative power the descriptor is able to obtain in the ideal case under our tests.
			Note that this is fair since all descriptors get the same treatment \cite{rfd}.

			We benchmark the retrieval performance with the nearest neighbor (NN), first tier (FT) and second tier (ST) scores.
			The idea is to check the ratio of retrieved objects in the query's class that also appear within the top $k$ matches.
			Specifically, for a class with $C$ members, $k=1$ for NN, $k=C-1$ for FT and $k=2(C-1)$ for ST.
			The final score is an average over all the objects in the database.

			Tables \ref{tbl:cmpfast} and \ref{tbl:cmpdog} show the retrieval results for our descriptors from Table \ref{tbl:skardescs}, our descriptor learned on the HPatches dataset \cite{hpatches} (included for completeness, see Section \ref{sec:hpatches} for details), recent state-of-the-art ones \cite{hardnet,convdesc,pnnet} and three "handcrafted" baselines (SIFT \cite{sift}, SURF \cite{surf} and intensity order features (IOF) from \cite{iof}, of which LIOP performed best and is included).
			\begin{table*}[htb]
	\centering
	\begin{subtable}[t]{1.6\columnwidth}
		\centering
		\resizebox{0.825\columnwidth}{!}
		{
\begin{tabular}{|c||c|c|c||c|c|c||c|c|c|}
	\hline
		\multirow{2}{*}{Descriptor}	& \multicolumn{3}{|c|}{UKB-test}	&	\multicolumn{3}{|c|}{ZuBuD}	&	\multicolumn{3}{|c|}{INRIA Holidays} \\
		\cline{2-10}
		&	\color{red}NN	&	\color{Aquamarine}FT	& \color{blue}ST	&	\color{red}NN	&	\color{Aquamarine}FT	& \color{blue}ST	&	\color{red}NN	&	\color{Aquamarine}FT	& \color{blue}ST	\\
	\hline
	\hline
	SKAR O-1	&	\color{red}$90.3$	&	\color{Aquamarine}$81.1$	&	\color{blue}$89.3$	&	\color{red}$94.5$	&	\color{Aquamarine}$79.2$	&	\color{blue}$86.2$	&	\color{red}$49.1$	&	\color{Aquamarine}$35.9$	&	\color{blue}$42.4$\\
	\hline
	SKAR O-6	&	\color{red}$95.9$	&	\color{Aquamarine}$88.0$	&	\color{blue}$93.1$	&	\color{red}$96.9$	&	\color{Aquamarine}$84.7$	&	\color{blue}$90.4$	&	\color{red}$54.6$	&	\color{Aquamarine}$40.1$	&	\color{blue}$47.1$\\
	\hline
	SKAR O-12	&	\color{red}$97.8$	&	\color{Aquamarine}$\bm{89.7}$	&	\color{blue}$\bm{93.5}$	&	\color{red}$96.7$	&	\color{Aquamarine}$85.8$	&	\color{blue}$91.1$	&	\color{red}$60.8$	&	\color{Aquamarine}$45.0$	&	\color{blue}$\bm{51.1}$\\
	\hline
	SKAR S-1	&	\color{red}$92.7$	&	\color{Aquamarine}$79.9$	&	\color{blue}$88.1$	&	\color{red}$91.2$	&	\color{Aquamarine}$75.1$	&	\color{blue}$82.0$	&	\color{red}$53.8$	&	\color{Aquamarine}$37.4$	&	\color{blue}$42.8$\\
	\hline
	SKAR S-6	&	\color{red}$93.5$	&	\color{Aquamarine}$81.2$	&	\color{blue}$87.9$	&	\color{red}$93.6$	&	\color{Aquamarine}$75.9$	&	\color{blue}$82.9$	&	\color{red}$58.2$	&	\color{Aquamarine}$41.0$	&	\color{blue}$47.0$\\
	\hline
	SKAR S-12	&	\color{red}$94.9$	&	\color{Aquamarine}$83.0$	&	\color{blue}$89.5$	&	\color{red}$94.7$	&	\color{Aquamarine}$77.8$	&	\color{blue}$83.9$	&	\color{red}$59.0$	&	\color{Aquamarine}$41.0$	&	\color{blue}$46.6$\\
	\hline
	SKAR S+O-12	&	\color{red}$\bm{98.0}$	&	\color{Aquamarine}$89.1$	&	\color{blue}$93.2$	&	\color{red}$96.9$	&	\color{Aquamarine}$85.5$	&	\color{blue}$90.8$	&	\color{red}$\bm{63.0}$	&	\color{Aquamarine}$\bm{45.3}$	&	\color{blue}$50.9$\\
	\hline
	SKAR-HPatches*	&	\color{red}$95.9$	&	\color{Aquamarine}${88.3}$	&	\color{blue}$93.2$	&	\color{red}$\bm{98.2}$	&	\color{Aquamarine}$\bm{88.5}$	&	\color{blue}$\bm{92.4}$	&	\color{red}$62.7$	&	\color{Aquamarine}$43.2$	&	\color{blue}$47.7$\\
	\hline
	DeepDesc \cite{convdesc}	&	\color{red}$72.2$	&	\color{Aquamarine}$57.6$	&	\color{blue}$68.7$	&	\color{red}$90.1$	&	\color{Aquamarine}$70.8$	&	\color{blue}$77.4$	&	\color{red}$41.2$	&	\color{Aquamarine}$26.8$	&	\color{blue}$31.8$\\
	\hline
	HardNet \cite{hardnet}	&	\color{red}$95.4$	&	\color{Aquamarine}$84.2$	&	\color{blue}$90.1$	&	\color{red}$97.4$	&	\color{Aquamarine}$85.0$	&	\color{blue}$89.5$	&	\color{red}$61.0$	&	\color{Aquamarine}$44.1$	&	\color{blue}$49.8$\\
	\hline
	PN-Net \cite{pnnet}	&	\color{red}$68.6$	&	\color{Aquamarine}$52.5$	&	\color{blue}$62.4$	&	\color{red}$88.5$	&	\color{Aquamarine}$66.7$	&	\color{blue}$73.8$	&	\color{red}$25.9$	&	\color{Aquamarine}$20.6$	&	\color{blue}$27.0$\\
	\hline
	SIFT \cite{sift}	&	\color{red}$74.2$	&	\color{Aquamarine}$51.2$	&	\color{blue}$58.3$	&	\color{red}$93.7$	&	\color{Aquamarine}$70.6$	&	\color{blue}$75.0$	&	\color{red}$21.7$	&	\color{Aquamarine}$17.3$	&	\color{blue}$23.4$\\
	\hline
	SURF \cite{surf}	&	\color{red}$66.0$	&	\color{Aquamarine}$47.1$	&	\color{blue}$56.9$	&	\color{red}$90.9$	&	\color{Aquamarine}$68.7$	&	\color{blue}$73.7$	&	\color{red}$24.6$	&	\color{Aquamarine}$17.8$	&	\color{blue}$23.3$\\
	\hline
	IOF \cite{iof}	&	\color{red}$51.6$	&	\color{Aquamarine}$35.3$	&	\color{blue}$42.5$	&	\color{red}$87.7$	&	\color{Aquamarine}$66.4$	&	\color{blue}$71.6$	&	\color{red}$20.1$	&	\color{Aquamarine}$12.5$	&	\color{blue}$15.3$\\
	\hline
\end{tabular}
		}
		\caption
		{
			Each image was represented with $\sim500$ ORB keypoints.
		}
		\label{tbl:cmpfast}
	\end{subtable}
	\begin{subtable}[t]{1.6\columnwidth}
		\centering
		\resizebox{0.825\columnwidth}{!}
		{
\begin{tabular}{|c||c|c|c||c|c|c||c|c|c|}
	\hline
		\multirow{2}{*}{Descriptor}	& \multicolumn{3}{|c|}{UKB-test}	&	\multicolumn{3}{|c|}{ZuBuD}	&	\multicolumn{3}{|c|}{INRIA Holidays} \\
		\cline{2-10}
		&	\color{red}NN	&	\color{Aquamarine}FT	& \color{blue}ST	&	\color{red}NN	&	\color{Aquamarine}FT	& \color{blue}ST	&	\color{red}NN	&	\color{Aquamarine}FT	& \color{blue}ST	\\
	\hline
	\hline
	SKAR O-1	&	\color{red}$38.7$	&	\color{Aquamarine}$31.0$	&	\color{blue}$43.0$	&	\color{red}$86.1$	&	\color{Aquamarine}$60.8$	&	\color{blue}$70.0$	&	\color{red}$12.0$	&	\color{Aquamarine}$8.9$	&	\color{blue}$12.5$\\
	\hline
	SKAR O-6	&	\color{red}$35.9$	&	\color{Aquamarine}$32.6$	&	\color{blue}$46.2$	&	\color{red}$86.3$	&	\color{Aquamarine}$63.3$	&	\color{blue}$72.3$	&	\color{red}$14.1$	&	\color{Aquamarine}$10.9$	&	\color{blue}$16.0$\\
	\hline
	SKAR O-12	&	\color{red}$60.5$	&	\color{Aquamarine}$47.9$	&	\color{blue}$59.5$	&	\color{red}$92.2$	&	\color{Aquamarine}$70.2$	&	\color{blue}$77.6$	&	\color{red}$17.6$	&	\color{Aquamarine}$13.9$	&	\color{blue}$19.1$\\
	\hline
	SKAR S-1	&	\color{red}$90.6$	&	\color{Aquamarine}$78.2$	&	\color{blue}$86.1$	&	\color{red}$96.4$	&	\color{Aquamarine}$80.0$	&	\color{blue}$85.9$	&	\color{red}$36.2$	&	\color{Aquamarine}$25.6$	&	\color{blue}$31.9$\\
	\hline
	SKAR S-6	&	\color{red}$94.8$	&	\color{Aquamarine}$84.6$	&	\color{blue}$89.5$	&	\color{red}$\bm{97.1}$	&	\color{Aquamarine}$81.3$	&	\color{blue}$86.6$	&	\color{red}$43.8$	&	\color{Aquamarine}$30.5$	&	\color{blue}$37.5$\\
	\hline
	SKAR S-12	&	\color{red}$94.2$	&	\color{Aquamarine}$84.5$	&	\color{blue}$89.4$	&	\color{red}$97.0$	&	\color{Aquamarine}$80.6$	&	\color{blue}$86.3$	&	\color{red}$43.0$	&	\color{Aquamarine}$30.4$	&	\color{blue}$36.9$\\
	\hline
	SKAR S+O-12	&	\color{red}$\bm{95.1}$	&	\color{Aquamarine}$\bm{84.8}$	&	\color{blue}$\bm{90.0}$	&	\color{red}$96.9$	&	\color{Aquamarine}$\bm{81.8}$	&	\color{blue}$\bm{86.9}$	&	\color{red}$\bm{46.8}$	&	\color{Aquamarine}$\bm{32.1}$	&	\color{blue}$\bm{37.9}$\\
	\hline
	SKAR-HPatches*	&	\color{red}$64.2$	&	\color{Aquamarine}$51.4$	&	\color{blue}$61.3$	&	\color{red}$93.7$	&	\color{Aquamarine}$73.7$	&	\color{blue}$77.8$	&	\color{red}$14.5$	&	\color{Aquamarine}$11.4$	&	\color{blue}$15.7$\\
	\hline
	DeepDesc \cite{convdesc}	&	\color{red}$38.5$	&	\color{Aquamarine}$30.5$	&	\color{blue}$40.9$	&	\color{red}$91.0$	&	\color{Aquamarine}$66.0$	&	\color{blue}$71.1$	&	\color{red}$5.5$	&	\color{Aquamarine}$4.4$	&	\color{blue}$7.1$\\
	\hline
	HardNet \cite{hardnet}	&	\color{red}$65.0$	&	\color{Aquamarine}$47.8$	&	\color{blue}$54.9$	&	\color{red}$92.1$	&	\color{Aquamarine}$69.2$	&	\color{blue}$73.6$	&	\color{red}$9.0$	&	\color{Aquamarine}$6.9$	&	\color{blue}$9.7$\\
	\hline
	PN-Net \cite{pnnet}	&	\color{red}$5.6$	&	\color{Aquamarine}$10.4$	&	\color{blue}$18.6$	&	\color{red}$80.6$	&	\color{Aquamarine}$51.2$	&	\color{blue}$56.8$	&	\color{red}$1.1$	&	\color{Aquamarine}$1.2$	&	\color{blue}$2.1$\\
	\hline
	SIFT \cite{sift}	&	\color{red}$80.6$	&	\color{Aquamarine}$67.1$	&	\color{blue}$75.8$	&	\color{red}$95.6$	&	\color{Aquamarine}$76.6$	&	\color{blue}$80.6$	&	\color{red}$23.8$	&	\color{Aquamarine}$21.9$	&	\color{blue}$29.9$\\
	\hline
	SURF \cite{surf}	&	\color{red}$7.6$	&	\color{Aquamarine}$10.1$	&	\color{blue}$18.5$	&	\color{red}$73.6$	&	\color{Aquamarine}$44.7$	&	\color{blue}$50.7$	&	\color{red}$0.3$	&	\color{Aquamarine}$0.4$	&	\color{blue}$0.7$\\
	\hline
	IOF \cite{iof}	&	\color{red}$29.4$	&	\color{Aquamarine}$22.7$	&	\color{blue}$30.6$	&	\color{red}$84.9$	&	\color{Aquamarine}$59.1$	&	\color{blue}$64.7$	&	\color{red}$7.6$	&	\color{Aquamarine}$6.6$	&	\color{blue}$10.4$\\
	\hline
\end{tabular}
		}
		\caption
		{
			Each image was represented with $\sim500$ SIFT (DoG) keypoints.
		}
		\label{tbl:cmpdog}
	\end{subtable}
	\caption
	{
		Retrieval results [\%] for different methods.
		First eight rows of both tables are for our descriptors.
	}
\end{table*}
			Important conclusions:
			\begin{enumerate}
				\item
					using augmented negative bags \eqref{eq:hardnegs} significantly improves SKAR descriptors;
				\item
					%tuning the descriptor for a particular keypoint type matters.
					tuning the SKAR descriptor to the dataset properties might matter quite a lot
					(e.g., to the used keypoint type).
			\end{enumerate}
			We can see from Table \ref{tbl:cmpfast} that our descriptors trained on ORB keypoints, SKAR O-1, O-6 and O-12, obtain excellent results for ORB keypoints.
			The SKAR descriptors learned on SIFT (DoG) keypoints, S-1, S-6 and S-12, do better than two baselines, SIFT and SURF, but are outperformed by HardNet \cite{hardnet}.
			The dataset based on SIFT keypoints (results in Table \ref{tbl:cmpdog}) is more difficult than the one based on ORB keypoints,
			as evidenced by worse performance of all descriptors\footnote{We double-checked our evaluation pipeline for corectness: some descriptors really do obtain such poor results in this setup. This may be due to the scaling parameters we used when cropping patches around SIFT keypoints.}.
			We conjecture this is due to the specific methodology we used to crop the patches around the detected keypoints
			(this methodology was initially chosen for ORB and left unchanged for SIFT).
			However, note that this is not important for our conclusions since all descriptors are tested on the same patches and we are interested in the relative ordering of retrieval scores (not their absolute values).
			On the experiments with SIFT keypoints (Table \ref{tbl:cmpdog}) we can see that SKAR descriptors learned on ORB keypoints exhibit poor retrieval results.
			However, SKAR descriptors learned on SIFT keypoints obtain best results\footnote{These descriptors (S-1, S-6 and S-12) also obtain solid results for ORB keypoints (Table \ref{tbl:cmpfast}) even though it can be seen as cross-dataset testing. We conjecture that this is due to the fact that our dataset of SIFT patches is a better training set due to its difficulty (results in better generalization).}.
			It is interesting to note that the SIFT descriptor outperforms other deep-learning approaches \cite{hardnet,convdesc,pnnet} for SIFT (DoG) keypoints.
			We can also see that using augmented negative bags \eqref{eq:hardnegs} during training improves the discriminative power of SKAR descriptors by a large margin, especially for keypoint types the descriptor was not trained on.
			Our descriptor learned on the HPatches dataset \cite{hpatches}, SKAR-HPatches*, obtains good results for ORB keypoints and is similar to HardNet \cite{hardnet} for SIFT keypoints.
			Our best descriptor, SKAR S+O-12, obtains excellent results in both setups.
			This indicates that the model from Table \ref{tbl:nn} has sufficient capacity to perform well for both ORB and SIFT (DoG) keypoints.
			However, it has too be tuned for the appearance of typical patches in the dataset
			(this is influenced, among other things, with the keypoint detector and its parameters).
			We suspect that all tested descriptors suffer from this issue\footnote{Rigorously validating this hypothesis is out of the scope of this paper.}.
			If true, the SKAR learning procedure is a good candidate to mitigate this since it enables learning from much weaker annotations than other approaches that require keypoint correspondences.

		\subsection{VLAD-based retrieval}
\begin{figure*}[ht]
				\centering
				\input{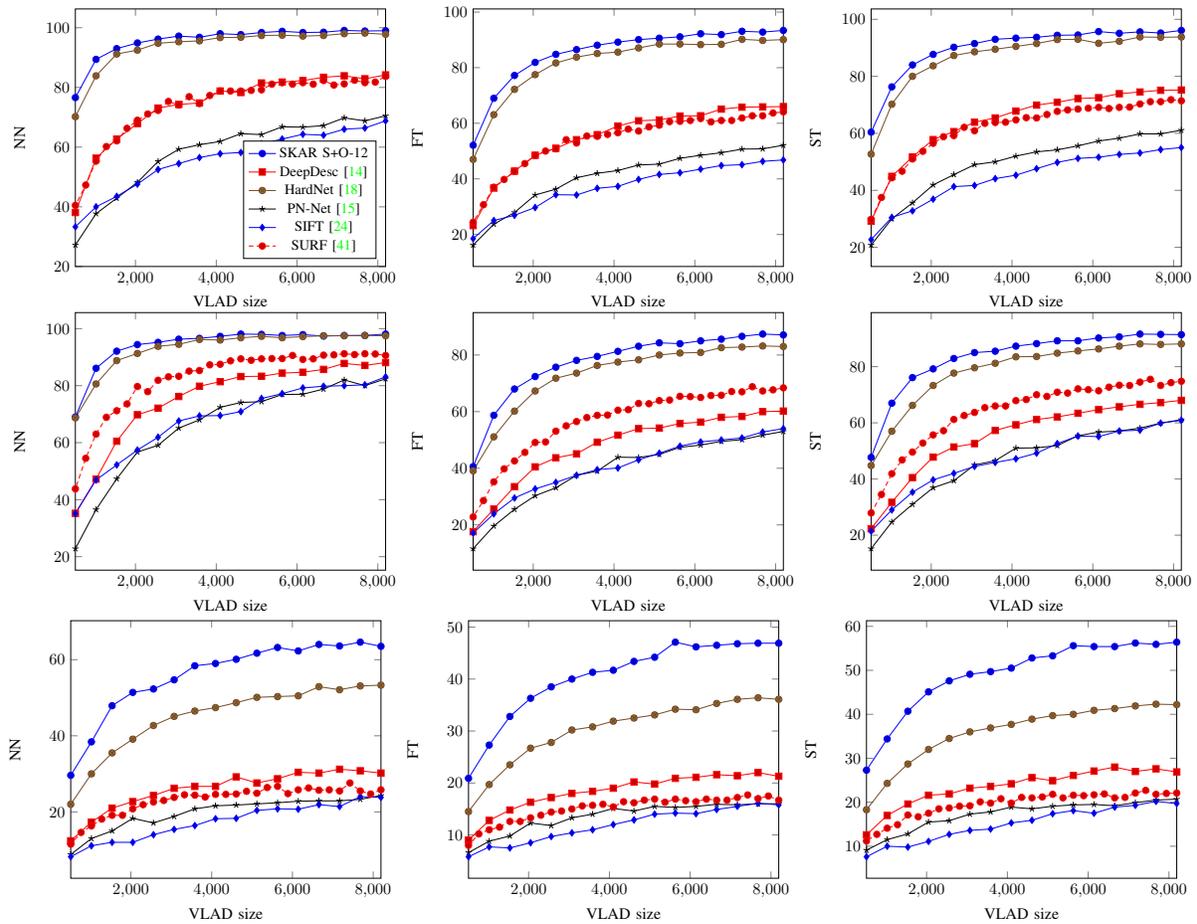}
				\caption
				{
					VLAD-based retrieval results for $\sim500$ ORB \cite{orb} keypoints per image on the UKB-test (first row), ZuBuD (second row) and INRIA Holidays (third row) datasets for varying number of centroids generated with $k$-means.
					The legend for all graphs is plotted in the top-left one.
					The VLAD size is the product of the local descriptor size and the number of centroids.
				}
				\label{fig:orbvlad}
			\end{figure*}

			In this subsection we experiment with image retrieval based on local feature aggregation.
			Note that none of the descriptors were tuned specifically for this task.
			For each image, we transform the extracted keypoints into descriptors and encode them with VLAD \cite{vlad} (a simplified Fisher kernel representation \cite{fisherkernel}).
			The centroids were generated with $k$-means on a subset of images.
			The similarity between two images is measured by an inner product between their VLADs.
			Figure \ref{fig:orbvlad} shows the NN, FT and ST VLAD retrieval scores on three datasets for different local descriptors when using $500$ ORB \cite{orb} keypoints per image on average.
			We included only our best descriptor, SKAR S+O-12, to reduce clutter.
			We see that this descriptor obtains excellent retrieval results.
			This indicates that the SKAR learning method is valuable even for non-matching tasks.

\begin{figure*}[ht]
				\centering
				\input{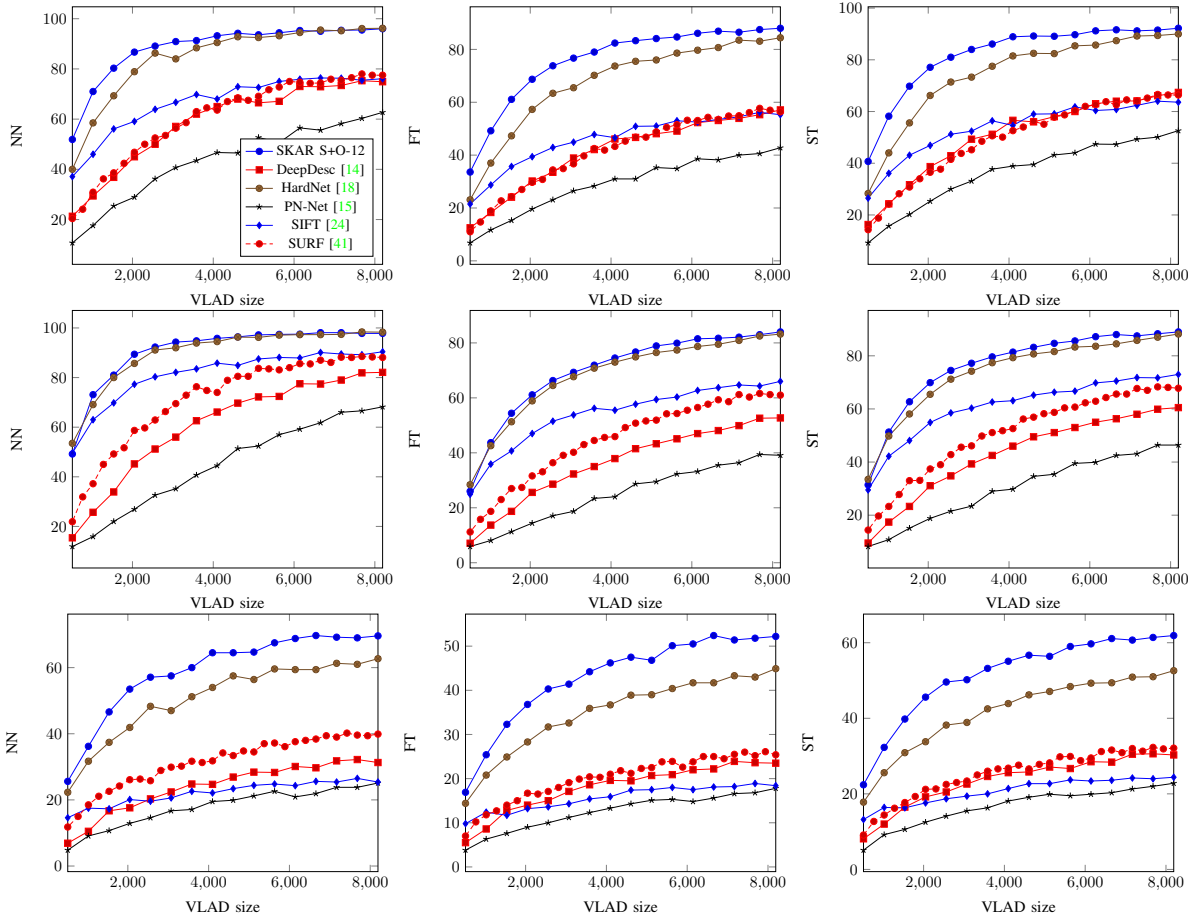}
				\caption
				{
					VLAD-based retrieval results for $\sim1000$ SURF \cite{surf} keypoints per image on the UKB-test (first row), ZuBuD (second row) and INRIA Holidays (third row) datasets.
					The graphs are of the same structure as those in Figure \ref{fig:orbvlad}.
				}
				\label{fig:surfvlad}
			\end{figure*}

			Figure \ref{fig:surfvlad} shows retrieval performance on SURF keypoints.
			Note that SKAR S+O-12 was not trained for their appearance.
			The results are similar to those presented in Figure \ref{fig:orbvlad}:
			HardNet \cite{hardnet} and our descriptor obtain similar performance on ZuBuD \cite{zubud} and UKB \cite{ukb} datasets,
			and our descriptor is the clear winner on INRIA Holidays \cite{holidays}.
			Other descriptors show weaker performance across all datasets.

			The main motivation for using our method is that it requires simpler training-set annotations.
			However, one might ask whether this really matters since annotated images abound nowadays and there is substantial experimental evidence that convolutional features are transferable (e.g., \cite{cnnofftheshelf}).
			We have already shown that descriptors learned on the dataset of Brown et al. \cite{brown} do not perform as good as our descriptors learned on the UKB dataset \cite{ukb}.
			We do not attribute this effect to the inherent superiority of our method or to the descriptor-extractor architecture we used in our experiments.
			We attribute this effect to the features of the training data:
			the patches extracted from the UKB images resemble testing data more than the patches obtained from the dataset of Brown et al. \cite{brown}.
			We claim that it is important to tune the parameters of the method to the task at hand
			(i.e., there are no completely transferable features).
			Note that this is a common opinion in the machine-learning community.
			Our experiments presented so far agree with this view.
			We provide further evidence in the next subsection with experiments in face verification.

		\subsection{Training descriptors for a specific task}
			Here we show that it is crucial to tune the descriptor for a specific task if high accuracy is desired.
			The tuning can be achieved with our method which requires simpler data annotation
			(potentially saves both time and money as opposed to standard descriptor-learning approaches).
			We show this through an experiment in face recognition/matching.

			We use the framework described by Li et al. \cite{porescale}.
			The basic idea is that the fine details on the skin of the face\footnote{These include face pores, fine wrinkles, hair, moles and small scars.} are unique for each individual and that this can be used for biometric applications.
			Given two face images, we determine whether or not they belong to the same person by matching these fine facial features.
			If we obtain a large number of matches, we can confidently claim that the images belong to the same person.
			Note that this approach is only applicable to high-resolution face images and that it differs from standard face-recognition methods (e.g., \cite{facenet}).

\begin{figure}
				\center
				\resizebox{0.5\columnwidth}{!}
				{
					\includegraphics[scale=1.0]{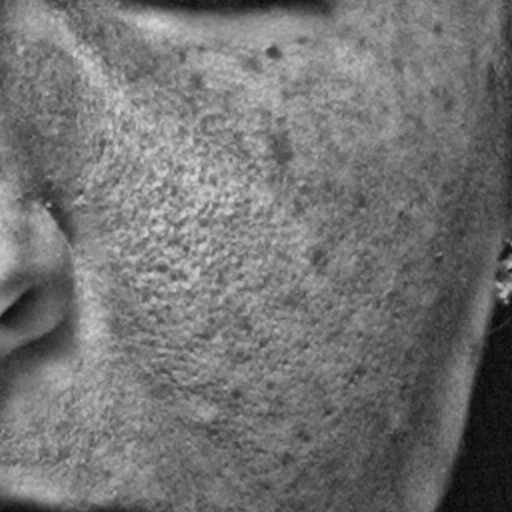}
				}
				\caption
				{
					For each face in the database \cite{bosphorus}, we crop a region under the left eye and use it in our experiments.
					Fine facial details are clearly visible after a contrast-normalization step.
				}
				\label{fig:bosphoruscrop}
			\end{figure}

			We use the Bosphorus database \cite{bosphorus} for our experiments.
			The database contains high-resolution face textures belonging to $105$ subjects.
			Multiple poses, expressions and occlusion conditions are present for each subject.
			In our experiment, we use only the frontal and near-frontal faces.
			We use the left cheek, i.e., the region of the face under the left eye, as a visual representation of the face.
			The region-cropping scheme can be seen in Figure \ref{fig:bosphoruscrop}.
			We use the SURF \cite{surf} keypoint detector to find $512$ keypoints within each region and extract $32\times 32$ patches around each of these keypoints.
			This pipeline transforms the cheek region into a bag of visual words.

\begin{figure*}
				\centering
				\resizebox{1.7\columnwidth}{!}
				{
					\input{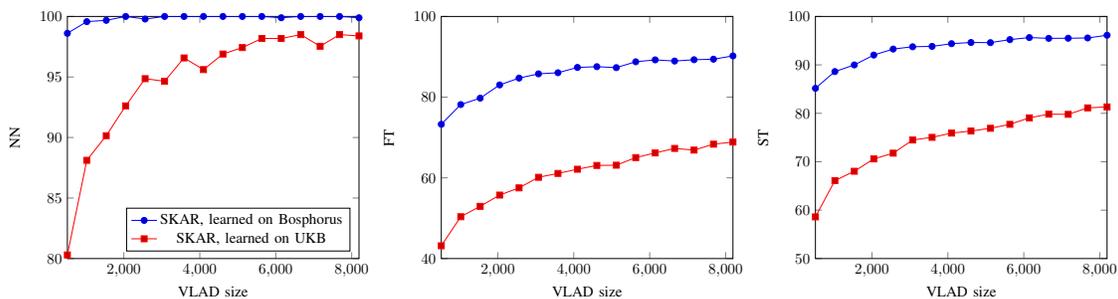}
				}
				\caption
				{
					The VLAD-based recognition accuracy for the testing subset of the Bosphorus dataset \cite{bosphorus}.
					The graphs are of the same structure as those in Figure \ref{fig:orbvlad}.
					The superior descriptor was learned on the training subset of the Bosphorus dataset.
				}
				\label{fig:bosphorusacc}
			\end{figure*}

			The obtained dataset is partitioned into a training, validation and testing subset.
			The training and validation subsets are used to learn a descriptor extractor with the same architecture as the one from our previous experiments (see Table \ref{tbl:nn}).
			We experimentally compare these two extractors on the generated testing subset.
			Note that both extractors have the same architectures.
			Figure \ref{fig:bosphorusacc} shows the VLAD-based recognition accuracy.
			We can see that learning the parameters for a particular task leads to large improvements.

			The next section demonstrates how the proposed method can be used to learn from unlabeled data.
			The experimental verification is done through learning convolutional features from unlabeled videos and retrieval of deformable 3D shapes.

	\section{Learning from unlabeled data}
		Unsupervised learning is a task of uncovering hidden structure from unlabeled data.
		The hope is that through this process one obtains useful information or features transferable to other tasks.
		Modern approaches that aim to learn convolutional features from unlabeled videos and images usually rely on a \textit{simple trick} that exploits the structure within these signals.
		Wang and Gupta \cite{wanggupta} use object tracking in videos:
		two patches connected by a track should have similar visual representations since they probably belong to same object or object part.
		Noroozi and Favaro \cite{noroozifavaro} partition an image into a $3\times3$ matrix of blocks, then randomly permute these blocks and learn a network to reassemble the original image.
		Misra et al. \cite{shuffleandlearn} use a similar idea:
		they shuffle several frames from a video and then learn a network to put them in a correct temporal order.
		Methods that learn by predicting the next sample in a sequence also exist \cite{pixelrnn,wavenet}.

		Here we describe two such simple tricks that enable learning of convolutional features from unlabeled data.
		We experimentally show that these features perform well on relevant tasks.
		The details are given in the following text.

		\subsection{Learning from unlabeled videos}\label{sec:unlabeledvid}
			Two frames that belong to the same video and are temporally close should have many matching keypoints.
			Also, two frames from unrelated videos should not have many matching keypoints.
			These trivial observations let us learn a discriminative local descriptor by using the method proposed in this paper.

			We perform experiments on the HUJI EgoSeg dataset \cite{egoseg}.
			The dataset contains approximately one hundred videos of people doing various activities:
			biking, running, cooking, sailing, driving, etc.
			We do not use any labels/annotations associated with the dataset.
			Also note that the public domain (YouTube) is full of such unlabeled videos.

			The training-set generation procedure proceeds as follows.
			For each video in the dataset, we extract several groups of frames separated apart by approximately $1$ minute
			(obviously, the number of groups depends on the length of the video).
			Each of these groups consists of $5$ frames spaced $250$ miliseconds apart.
			Each frame is transformed into a bag of $512$ keypoints with the SURF detector.
			Two bags that come from the same group form a positive pair
			(i.e., we assume that they have many matching keypoints).
			These form a triplet (Equation \ref{eq:bagtriplets}) with any bag of keypoints that comes from some other group of frames.
			The described procedure enables us to generate a large training dataset for our method.
			We use this dataset to learn the model specified in Table \ref{tbl:nn}.
			We compare this model to the one learned on the UKB training images \cite{wlrnicpr}
			(same architecture, also trained on $512$ SURF keypoints per bag).

\begin{figure*}[ht]
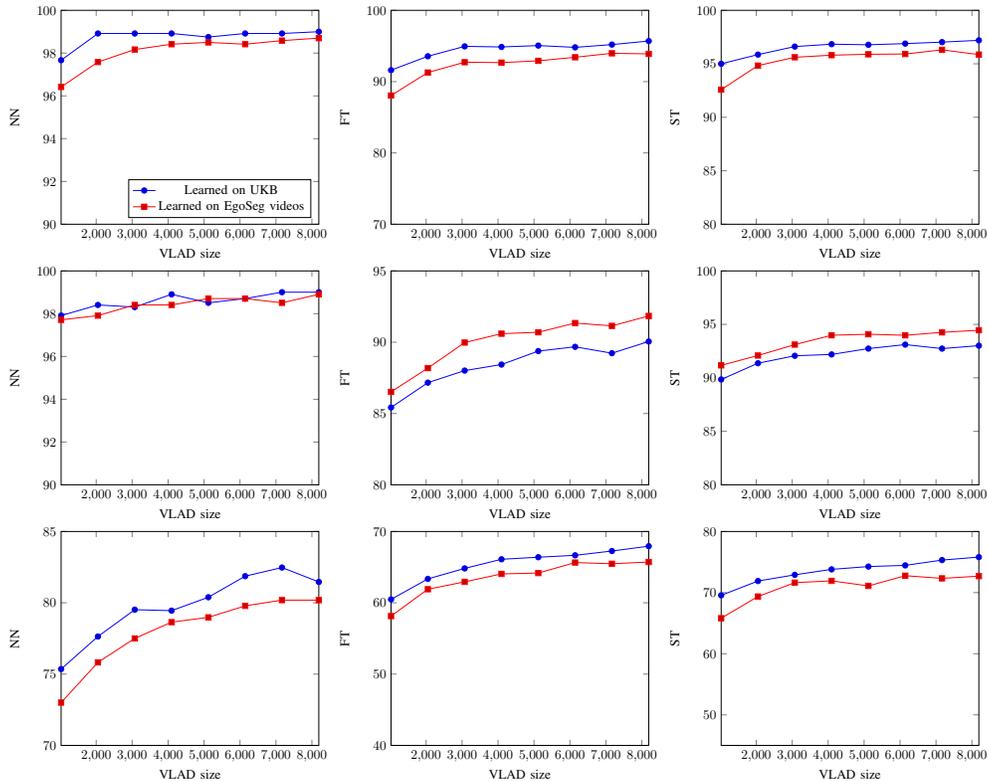

				\centering
				\resizebox{1.5\columnwidth}{!}
				{
					\input{ukb-vlad2.plot}
				}
				\resizebox{1.5\columnwidth}{!}
				{
					\input{zubud-vlad2.plot}
				}
				\resizebox{1.5\columnwidth}{!}
				{
					\input{holidays-vlad2.plot}
				}
				\caption
				{
					VLAD-based retrieval results on the UKB-test (first row), ZuBuD (second row) and INRIA Holidays (third row).
					The graphs are of the same structure as those in Figure \ref{fig:orbvlad}.
					We can see that both models perform approximately the same.
				}
				\label{fig:vlad2}
			\end{figure*}

			Figure \ref{fig:vlad2} shows VLAD-based retrieval results.
			We can see that both models perform approximately the same.
			This is a confirmation that our approach enables learning of useful convolutional features from unlabeled videos.
			%Next, we describe how a similar idea can be used for the retrieval of 3D shapes.
			Next, we apply similar ideas to the retrieval of 3D shapes.

		\subsection{Learning 3D-shape retrieval from unlabeled data}
			Our plan is to learn a shape-retrieval system from labels (annotations) generated in an automatic way.
			To achieve this in our framework, we transform the problem into an image-retrieval task by rendering each 3D shape from multiple views \cite{lightfield} and transforming the resulting images into bags of keypoints.
			Note that some bag pairs are expected to have many matching keypoints between them.
			Specifically, those that are extracted from related views of the same 3D shape.
			On the other hand, bag pairs that come from two unrelated shapes should not have many matching keypoints.
			These observations enable us to generate a dataset for our method (Equation \eqref{eq:bagtriplets}) without any labels associated with each 3D shape.

\begin{figure}
				\centering

				\resizebox{0.45\columnwidth}{!}
				{
					\includegraphics[scale=0.12]{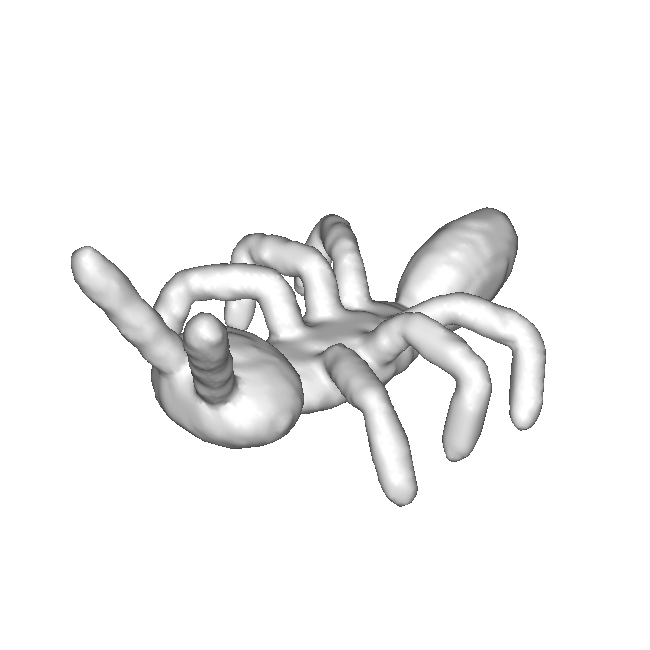}
					\includegraphics[scale=0.12]{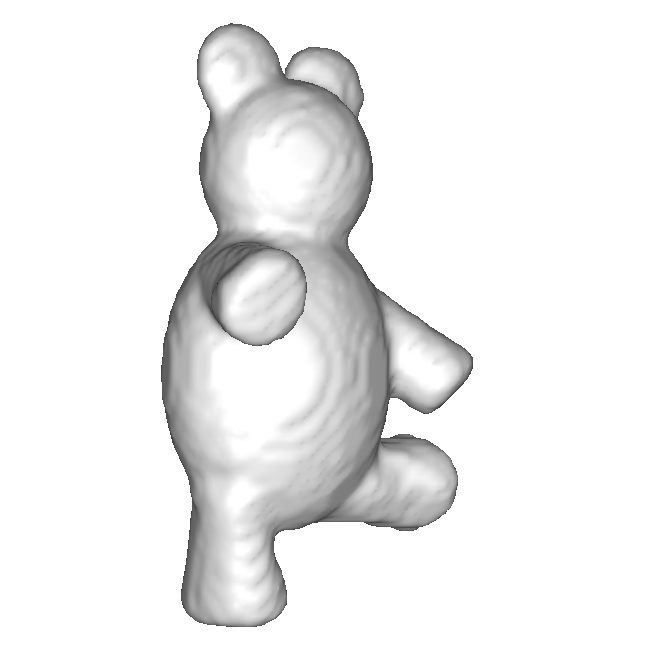}
				}
				\resizebox{0.45\columnwidth}{!}
				{
					\includegraphics[scale=0.12]{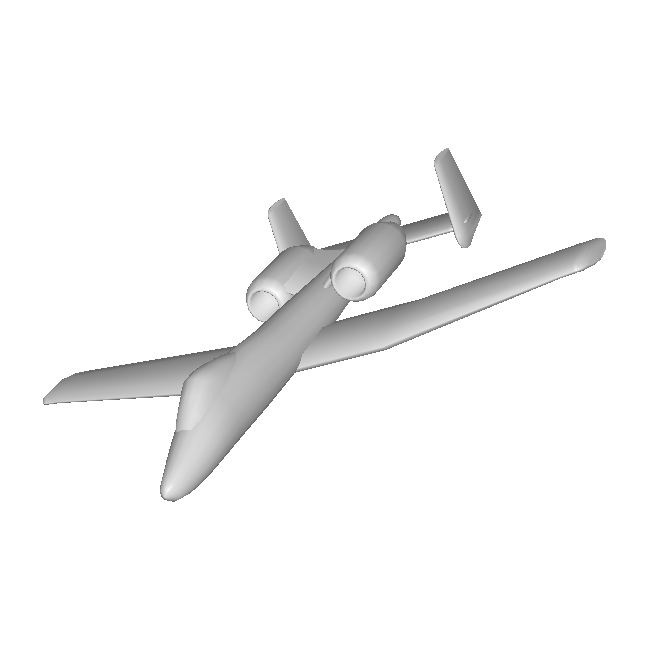}
					\includegraphics[scale=0.12]{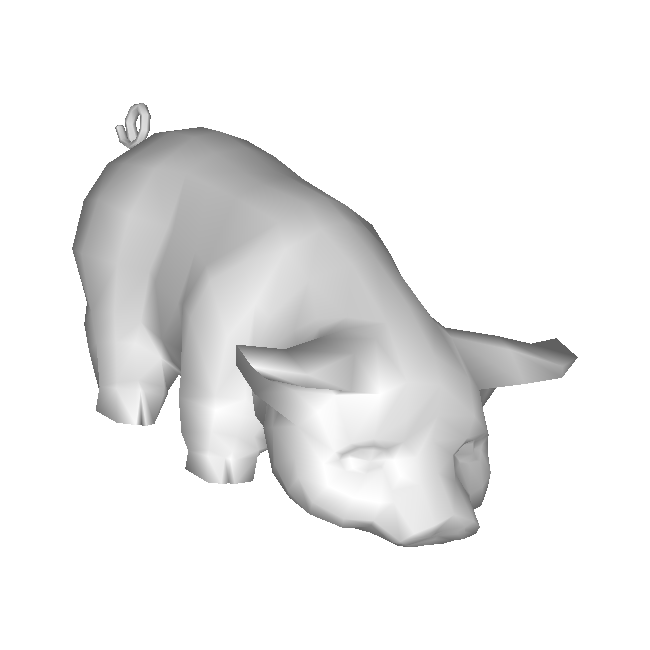}
				}
				\caption
				{
					Samples from the McGill \cite{mcgill} (first two shapes) and PSB \cite{psb} (second two shapes) 3D-shape datasets.
				}
				\label{fig:3d}
			\end{figure}

			We use the McGill \cite{mcgill} and PSB \cite{psb} datasets for our experiments.
			The McGill dataset contains $255$ shapes with significant part articulations grouped into 10 classes
			(ants, spiders, crabs, humans, etc.).
			The PSB dataset is larger:
			it contains 1814 shapes grouped into 90 classes.
			Examples from both datasets can bee seen in Figure \ref{fig:3d}.

			We generate the training data from the PSB dataset.
			We achieve this by rendering $4$ groups of $32$ randomly chosen views for each shape.
			Each of these groups is transformed into a bag of keypoints by keeping $512$ most salient ORB keypoints \cite{orb} extracted from the $32$ views belonging to the group.
			This procedure results in $4\times 1814$ keypoint bags.
			Two bags form a matching pair if they were generated from the same shape.
			Together with some unrelated bag, they form a triplet from Equation \eqref{eq:bagtriplets}.
			We use this data to learn a model with architecture specified in Table \ref{tbl:nn}.

			To test the method, we use the McGill dataset.
			Each shape is transformed into a keypoint bag with the similar procedure as the one used to generate the training data.
			The difference is that we render just one group of $32$ views per shape (instead of $4$).
			Consequently, there is just one bag of keypoints for each shape.
			We use this approach to adhere to the standard testing protocol on the McGill dataset.
			Table \ref{tbl:mcgill} compares VLAD-based retrieval scores ($64$ centroids) achieved by our approach to some other methods from the literature.
			\begin{table}
				\centering
				\resizebox{0.7\columnwidth}{!}
				{
					\begin{tabular}{|c||c|c|c|}
						\hline
						Method	&	NN	&	FT	&	ST	\\
						\hline
						\hline
						SKAR (unlabeled)	&	$96.8$	&	$73.9$	&	$90.4$	\\
						\hline
						GIFT \cite{gift}	&	$98.4$	&	$90.5$	&	$97.3$	\\
						\hline
						DeepShape \cite{dasd}	&	$98.8$	&	$78.2$	&	$83.4$	\\
						\hline
						Covariance descriptors \cite{cov3d}	&	$97.7$	&	$73.2$	&	$81.8$	\\
						\hline
						Graph-based \cite{graph3d}	&	$97.6$	&	$74.1$	&	$91.1$	\\
						\hline
						3D SIFT \cite{sift3d}	&	$97.2$	&	$65.8$	&	$78.4$	\\
						\hline
						VLAT \cite{vlat}	&	$96.9$	&	$65.8$	&	$78.1$	\\
						\hline
						Hybrid BOW \cite{hybridbow}	&	$95.7$	&	$63.5$	&	$79.0$	\\
						\hline
						Hybrid 2D/3D \cite{hybrid2d3d}	&	$92.5$	&	$55.7$	&	$69.8$	\\
						\hline
					\end{tabular}
				}
				\caption
				{
					Retrieval performance [\%] on the McGill dataset.
				}
				\label{tbl:mcgill}
			\end{table}
			We can see that our approach obtains comparable scores to all methods except to the ones based on CNNs learned in a supervised way directly for shape retrieval \cite{dasd,gift}.
			However, one should note that the DeepShape method \cite{dasd} used a part of the McGill dataset ($\approx40$\%) for learning and reported retrieval scores on the rest:
			not all subsets are of the same difficulty so their results are ambiguous.
			The GIFT \cite{gift} system involves a re-ranking component \cite{queryrankfusion} among other augmentations.
			Our retrieval pipeline would also benefit from these.
			However, our main goal was to show that useful convolutional features can be learned from unlabeled 3D shapes:
			the features perform as well or better than other handcrafted features.

	\section{Experiments on the HPatches benchmark}\label{sec:hpatches}
		The experiments presented so far have all been designed by us.
		To have a more fair and independent comparison, here we present experiments on the HPatches benchmark \cite{hpatches}.
		This benchmark enables an objective comparison of local image descriptors on a large dataset.
		The images in the dataset contain significant illumination and viewpoint changes.
		The matches between local keypoints in the corresponding images are provided as ground truth.
		To simulate the noise that the keypoint detectors introduce in practice, the precomputed keypoints are perturbed in their position, scale and orientation by three increasing noise levels:
		easy, hard and tough.

		The evaluation is done through a strict protocol consisting of three tasks: matching, retrieval and verification.
		The performance on each of the tasks provides an insight into the descriptor's potential for a certain application
		(the tasks were designed to imitate typical use cases).
		The matching task measures how many keypoints does a descriptor match correctly between a target and a reference image.
		The retrieval task measures how well a descriptor retrieves similar patches from a large collection.
		The verification task measures how well a descriptor separates positive from negative pairs of patches.
		The performance on each of the tasks is measured by precision/recall and their variations.
		For more details, see the paper that introduced the benchmark \cite{hpatches}.

		For our experiments, we use the same training/testing data partition as the one used in the ECCV2016 workshop "Local Features: State of the art, open problems and performance evaluation"
		(this is split "a" in the HPatches benchmark).
	
		We report the results for four descriptors learned with our method, all having the same architecture.
		The descriptors SKAR-EgoSeg and SKAR-EgoSeg* were learned on the EgoSeg dataset (see Section \ref{sec:unlabeledvid} for details) without and with mining hard negatives, respectively.
		The descriptors SKAR-HPatches and SKAR-HPatches* were learned on the training partition of the HPatches dataset\footnote{We used only image-level annotations when constructing keypoint bags. Also, note that the training and testing sequences are disjoint, so the keypoints on which the evaluation is performed are not included in the training set.} without and with mining hard negatives, respectively.
		The augmented negative bags $K^-_*$ (Equation \eqref{eq:hardnegs}) were obtained by merging $6$ negative bags.
		Our descriptors are compared to SIFT \cite{sift} and four convolutional descriptors \cite{tfeat,zagoruyko,convdesc,hardnet} learned on the dataset of Borwn et al. \cite{brown}.
		See tables \ref{tbl:hpatches-verification-inter}, \ref{tbl:hpatches-verification-intra}, \ref{tbl:hpatches-matching} and \ref{tbl:hpatches-retrieval} for the results on each of the tasks.
		\begin{table*}
			\begin{subtable}{.5\linewidth}
				\centering
				\begin{tabular}{|c||c|c|c|}
					\hline
					Noise level	&	Easy	&	Hard	&	Tough	\\
					\hline
					\hline
					SIFT \cite{sift}	&	$0.849$	&	$0.657$	&	$0.512$	\\
					\hline
					TFeat \cite{tfeat}	&	$0.912$	&	$0.848$	&	$0.752$	\\
					\hline
					DeepCompare \cite{zagoruyko}	&	$0.845$	&	$0.724$	&	$0.609$	\\
					\hline
					DeepDesc \cite{convdesc}	&	$0.904$	&	$0.830$	&	$0.733$	\\
					\hline
					HardNet \cite{hardnet}	&	$0.949$	&	$0.901$	&	$\mathbf{0.811}$	\\
					\hline
					\hline
					SKAR-EgoSeg	&	$0.896$	&	$0.747$	&	$0.609$	\\
					\hline
					SKAR-EgoSeg*	&	$0.903$	&	$0.770$	&	$0.637$	\\
					\hline
					\hline
					SKAR-HPatches	&	$0.949$	&	$0.900$	&	$0.809$	\\
					\hline
					SKAR-HPatches*	&	$\mathbf{0.950}$	&	$\mathbf{0.902}$	&	$0.806$	\\
					\hline
				\end{tabular}
				\caption
				{
					Negative patch pairs sampled from different sequences.
				}
				\label{tbl:hpatches-verification-inter}
			\end{subtable}% <--- vazan je "%"
			\begin{subtable}{.5\linewidth}
				\centering
				\begin{tabular}{|c||c|c|c|}
					\hline
					Noise level	&	Easy	&	Hard	&	Tough	\\
					\hline
					\hline
					SIFT \cite{sift}	&	$0.783$	&	$0.570$	&	$0.429$	\\
					\hline
					TFeat \cite{tfeat}	&	$0.868$	&	$0.781$	&	$0.671$	\\
					\hline
					DeepCompare \cite{zagoruyko}	&	$0.785$	&	$0.644$	&	$0.524$	\\
					\hline
					DeepDesc \cite{convdesc}	&	$0.851$	&	$0.751$	&	$0.640$	\\
					\hline
					HardNet \cite{hardnet}	&	$0.923$	&	$0.858$	&	$\mathbf{0.752}$	\\
					\hline
					\hline
					SKAR-EgoSeg	&	$0.852$	&	$0.678$	&	$0.532$	\\
					\hline
					SKAR-EgoSeg*	&	$0.866$	&	$0.713$	&	$0.572$	\\
					\hline
					\hline
					SKAR-HPatches	&	$0.919$	&	$0.848$	&	$0.734$	\\
					\hline
					SKAR-HPatches*	&	$\mathbf{0.927}$	&	$\mathbf{0.863}$	&	$0.751$	\\
					\hline
				\end{tabular}
				\caption
				{
					Negative patch pairs sampled from the same sequence.
				}
				\label{tbl:hpatches-verification-intra}
			\end{subtable}
			\caption
			{
				Average-precision scores on the verification task of the HPatches benchmark.
				Each experiment consists of $2\times 10^5$ positive pairs and $10^6$ negative pairs.
			}
		\end{table*}
		\begin{table}
			\centering
			\begin{tabular}{|c||c|c|c|}
				\hline
				Noise level	&	Easy	&	Hard	&	Tough	\\
				\hline
				\hline
				SIFT \cite{sift}	&	$0.453$	&	$0.193$	&	$0.086$	\\
				\hline
				TFeat \cite{tfeat}	&	$0.467$	&	$0.259$	&	$0.131$	\\
				\hline
				DeepCompare \cite{zagoruyko}	&	$0.399$	&	$0.199$	&	$0.095$	\\
				\hline
				DeepDesc \cite{convdesc}	&	$0.430$	&	$0.239$	&	$0.125$	\\
				\hline
				HardNet \cite{hardnet}	&	$0.658$	&	$0.479$	&	$0.303$	\\
				\hline
				\hline
				SKAR-EgoSeg	&	$0.495$	&	$0.226$	&	$0.102$	\\
				\hline
				SKAR-EgoSeg*	&	$0.507$	&	$0.251$	&	$0.123$	\\
				\hline
				\hline
				SKAR-HPatches	&	$0.677$	&	$0.494$	&	$0.313$	\\
				\hline
				SKAR-HPatches*	&	$\mathbf{0.695}$	&	$\mathbf{0.525}$	&	$\mathbf{0.342}$	\\
				\hline
			\end{tabular}
			\caption
			{
				Mean average precision on the image-matching task of the HPatches benchmark, measured over multiple predefined image pairs.
			}
			\label{tbl:hpatches-matching}
		\end{table}
		\begin{table}
			\centering
			\begin{tabular}{|c||c|c|c|}
				\hline
				Noise level	&	Easy	&	Hard	&	Tough	\\
				\hline
				\hline
				SIFT \cite{sift}	&	$0.545$	&	$0.264$	&	$0.134$	\\
				\hline
				TFeat \cite{tfeat}	&	$0.551$	&	$0.344$	&	$0.189$	\\
				\hline
				DeepCompare \cite{zagoruyko}	&	$0.527$	&	$0.309$	&	$0.170$	\\
				\hline
				DeepDesc \cite{convdesc}	&	$0.561$	&	$0.374$	&	$0.228$	\\
				\hline
				HardNet	\cite{hardnet}	&	$0.725$	&	$0.572$	&	$0.385$	\\
				\hline
				\hline
				SKAR-EgoSeg	&	$0.595$	&	$0.319$	&	$0.169$	\\
				\hline
				SKAR-EgoSeg*	&	$0.605$	&	$0.344$	&	$0.189$	\\
				\hline
				\hline
				SKAR-HPatches	&	$0.756$	&	$0.621$	&	$0.440$	\\
				\hline
				SKAR-HPatches*	&	$\mathbf{0.770}$	&	$\mathbf{0.645}$	&	$\mathbf{0.464}$	\\
				\hline
			\end{tabular}
			\caption
			{
				Mean average precision on the retrieval task of the Hpatches benchmark
				($10^4$ experiemnts of retrieving $5$ patches corresponding to a query among $2\times 10^4$ distractors).
			}
			\label{tbl:hpatches-retrieval}
		\end{table}

		The descriptors learned with our method on the HPatches training dataset, SKAR-HPatches and SKAR-HPatches*, obtain very good results when compared to competing approaches,
		especially on the image-matching task for which they were explicitly tuned for.
		By comparison, the SKAR-EgoSeg descriptor obtains significantly worse results on all three tasks, especially in the "hard" and "tough" noise-level settings.
		This is despite having the same architecture as SKAR-HPatches.
		However, note that SKAR-EgoSeg was learned on SURF keypoints extracted from a dataset unrelated to HPatches testing data which consists of DoG and Hessian keypoints.
		This provides more evidence that it is crucial to tune the descriptor for the visual appearance and properties of the testing images.
		Hard-negative mining helps significantly:
		both SKAR-EgoSeg* and SKAR-Hpatches* outperform their counterparts learned with our original method \cite{wlrnicpr}.
		Bearing in mind that hard negatives can be easily obtained most of the time, i.e., without any additional data-labeling efforts,
		it is useful to include the proposed mining procedure in the learning loop since it boosts performance.

		As for other convolutional descriptors, TFeat \cite{tfeat}, DeepCompare \cite{zagoruyko} and DeepDesc \cite{convdesc} obtain similar results to SKAR-EgoSeg and outperform SIFT \cite{sift}
		(although not by a large margin).
		The HardNet descriptor \cite{hardnet}, which uses the L2-Net architecture \cite{l2net} but improves on its learning procedure,
		obtains strong results, especially on the verification task.
		We conjecture that high performance on this task is due to the learning procedure that uses data in the form of keypoint triplets with hard-negative mining.
		However, our strongest descriptor, SKAR-HPatches*, clearly outperforms HardNet on the matching and retrieval tasks.
		This is despite the fact that the neural network used by our descriptors has $5$ times less parameters and being $4$ times faster when processing a patch than L2-Net \cite{l2net} used by HardNet.
		This demonstrates that it is more important to learn the network for a specific appearance of the dataset than increasing its size for performance gains.
		Our method makes this task easier to achieve since it requires smaller dataset-annotation efforts than traditional descriptor-learning approaches.

	\section{Conclusion}
		We point out that the current best methods for learning local descriptors require a large number of matching and non-matching keypoint pairs.
		Data of this kind is not always available and, thus, these methods are not always applicable.
		To address this issue, we introduce and analyze an algorithm for learning local descriptors from weakly-labeled datasets and discuss the improvements that could be obtained through the process of hard-negative mining.
		The experiments show that our descriptors compare well to the best available ones and significantly outperform SIFT.
		We also show how to learn useful convolutional features from unlabeled videos and 3D shapes.
		Note that these properties of our method enable the learning of local descriptors from datasets with much simpler annotations (or none at all).
		This potentially saves time and reduces cost when building a computer-vision system that relies on highly discriminative local descriptors.
		The code is available online as a Git repository
		(use commit \texttt{ab4392d4eb87c25c349d2d6ffb514bee2860f8f0}).

	\small
	{
		\textbf{Acknowledgements.}
		This research was partially supported by Visage Technologies AB (Link\"oping, Sweden), and by the Croatian Science Foundation (project 8065).
	}

\ifarxiv
	\balance
\else
	
\fi

	\bibliographystyle{IEEEtran}
	\bibliography{references}

\ifarxiv
	
\else
	\input{biographies/markus.tex}
	\input{biographies/pandzic.tex}
	\input{biographies/ahlberg.tex}
\fi

\end{document}